\documentclass[conference]{IEEEtran}
\IEEEoverridecommandlockouts
\usepackage{cite}
\usepackage{amsmath,amssymb,amsfonts}
\usepackage{algorithmic}
\usepackage{graphicx}
\usepackage{textcomp}
\usepackage{xcolor, hyperref, todonotes, booktabs, multicol, multirow, listings, subcaption, orcidlink}
\def\BibTeX{{\rm B\kern-.05em{\sc i\kern-.025em b}\kern-.08em
    T\kern-.1667em\lower.7ex\hbox{E}\kern-.125emX}}

\renewcommand\thefigure{\arabic{figure}}
\newcommand{\coo}{\ensuremath{\mathrm{CO_2}}}

\lstdefinestyle{python}{
    backgroundcolor=\color{backcolour},   
    commentstyle=\color{codegreen},
    keywordstyle=\color{magenta},
    numberstyle=\tiny\color{codegray},
    stringstyle=\color{codepurple},
    basicstyle=\ttfamily\footnotesize,
    breakatwhitespace=false,         
    breaklines=true,                 
    captionpos=b,                    
    keepspaces=true,                 
    numbers=left,                    
    numbersep=5pt,                  
    showspaces=false,                
    showstringspaces=false,
    showtabs=false,                  
    tabsize=2
}
\lstdefinestyle{json}{
    basicstyle=\footnotesize\ttfamily,
    numbers=left,
    numberstyle=\tiny,
    stepnumber=1,
    numbersep=5pt,
    backgroundcolor=\color{gray!10},
    showstringspaces=false,
    breaklines=true,
    frame=single,
    captionpos=b,
    keywordstyle=\color{blue},
    commentstyle=\color{gray},
    stringstyle=\color{red},
    literate=
     *{0}{{{\color{blue}0}}}{1}
      {1}{{{\color{blue}1}}}{1}
      {2}{{{\color{blue}2}}}{1}
      {3}{{{\color{blue}3}}}{1}
      {4}{{{\color{blue}4}}}{1}
      {5}{{{\color{blue}5}}}{1}
      {6}{{{\color{blue}6}}}{1}
      {7}{{{\color{blue}7}}}{1}
      {8}{{{\color{blue}8}}}{1}
      {9}{{{\color{blue}9}}}{1}
      {:}{{{\color{red}{:}}}}{1}
      {,}{{{\color{red}{,}}}}{1}
      {\{}{{{\color{blue}{\{}}}}{1}
      {\}}{{{\color{blue}{\}}}}}{1}
      {[}{{{\color{blue}{[}}}}{1}
      {]}{{{\color{blue}{]}}}}{1}
}
\begin{document}

\title{An exploration of the effect of quantisation on energy consumption and inference time of StarCoder2}



\author{\IEEEauthorblockN{Pepijn de Reus}
\IEEEauthorblockA{\textit{Master Artificial Intelligence} \\
\textit{University of Amsterdam}\\
Amsterdam, The Netherlands \\
p.dereus@uva.nl}
\and
\IEEEauthorblockN{Ana Oprescu}
\IEEEauthorblockA{\textit{Complex Cyber Infrastructure} \\
\textit{University of Amsterdam}\\
Amsterdam, The Netherlands \\
a.m.oprescu@uva.nl}
\and
\IEEEauthorblockN{Jelle Zuidema}
\IEEEauthorblockA{\textit{Institute for Language, Logic and Computation} \\
\textit{University of Amsterdam}\\
Amsterdam, The Netherlands \\
w.h.zuidema@uva.nl}
}

\maketitle

\begin{abstract}
This study examines quantisation and pruning strategies to reduce energy consumption in code Large Language Models (LLMs) inference. Using StarCoder2, we observe increased energy demands with quantization due to lower throughput and some accuracy losses. Conversely, pruning reduces energy usage but impairs performance. The results highlight challenges and trade-offs in LLM model compression. We suggest future work on hardware-optimized quantization to enhance efficiency with minimal loss in accuracy.
\end{abstract}


\section{Introduction}
In recent years, Artificial Intelligence (AI) has gained unprecedented attention from the scientific community and the general public. With the introduction of ChatGPT, using Large Language Models
(LLMs) became accessible for those without coding experience. As a result, ChatGPT gained over 180 million users in the first year~\cite{ghassemi_chatgpt_2023}. This popularity spread to software engineers as well, where 83\% use AI to code~\cite{yepis_hype_2024}. One of the biggest code-sharing platforms, GitHub, deployed its code LLM called GitHub Copilot in 2021. GitHub Copilot's users report increased satisfaction and productivity when coding with the model~\cite{kalliamvakou_quantifying_2022}.

Besides these benefits, these developments also bring challenges. Since the wide adoption of LLMs as Chat-GPT, reviews, theses and even scientific articles have been written by LLMs. Current research often focuses on mitigating the risks of truthfulness and interpretability, but little work is conducted on the energy consumption of LLMs. Increasingly, scientists call to reduce the energy consumption of AI, as the use of AI significantly impacts global energy consumption~\cite{de_vries_growing_2023}. Alphabet reported that its energy bill increased over tenfold due to the required computational power to train AI models, and both Microsoft and Google reported an increased carbon emissions. To a large extent, this can be attributed to the size of the models and data that keeps increasing~\cite{han_pre-trained_2021}. Where literature often studies the energy consumption of training a model~\cite{verdecchia_systematic_2023}, the inference stage, where users interact with the model, is often neglected. However, as the popularity of these applications increases, so does the number of queries for the model. Experts estimate that three days of ChatGPT consumes enough energy to match the training phase~\cite{de_vries_growing_2023}. Thus, the inference phase is where the most impact can be achieved in reducing energy consumption. Moreover, this phase is to a larger extent under user control compared to the training phase. To run these large models on small devices such as laptops (or even phones), users can shrink the model with quantisation~\cite{almeida_smart_2021}. Model weights are then stored in a lower-bit format to reduce memory.

Combining these trends, we identify an interesting area for exploration, and formulate the following research questions: \\
\textbf{RQ1.} How can we reduce the energy consumption of a code LLM using quantisation with minimal harm to accuracy?\\
\textbf{RQ2.} How can we reduce the energy consumption of a code LLM using pruning with minimal harm to accuracy?

We present the foundation and relevant theoretical background in Section~\ref{SecTheBack}, then our methodology in Section~\ref{SecMet} and the experimental setup inSection~\ref{SecExpSet}, followed by the results in Section~\ref{SecRes}. We interpret and discuss the results in Section~\ref{SecDis}, where we also state our limitations and provide directions for future work. We then provide an overview of similar research in Section~\ref{SecRelWork}. Finally, we summarise our exploration in Section~\ref{SecCon}.

\section{Background} \label{SecTheBack}

\subsection{Global warming}
In 2023, global temperatures for the first time exceeded the 1.5 Celsius warming limit of the Paris Agreement for consecutive months~\cite{copernicus_global_2024}. Overall, global average temperatures were 1.48 degrees Celsius above pre-industrial, leaving a fragile margin to remain within the Paris Agreement. This makes it likely that 1.5 degrees Celsius warming will not be achieved in 2050, as per the Paris Agreement, but already in this decade with 80\% probability~\cite{wmo_world_2024, hansen_global_2023}. Already this causes extreme heatwaves, which are now 35 times more likely than before the industrial age~\cite{pinto_extreme_2024}. This global warming is then expected to exceed 2 Celsius before 2050~\cite{hansen_global_2023}, increasing the likelihood of irreversible changes such as large-scale death of coral reefs and permanent melting of the Arctic sea~\cite{lenton_climate_2019}. Rising energy consumption and rising global temperatures go hand-in-hand, so there is an urgent demand to reduce the carbon emissions of AI.

\subsection{Language models writing code}
There is a long history of language models, to keep this subsection compact we will stick to the most recent developments relevant to the domain of code LLMs. The first LLM was ELMo~\cite{peters_deep_2018}, a biLSTM~\cite{sutskever_sequence_2014} architecture that allows taking context into account because it processes language both forward and backward. Together with it's deep contextualized representations, ELMo allows more context into the prediction. Until 2015 the field of neural machine translation used an encoder-decoder architecture based on a Recurrent Neural Network (RNN) to translate a source sentence~\cite{bahdanau_neural_2016}. The encoder neural network encodes a source sentence into a fixed-length vector, the decoder outputs a translation of the encoded fixed-length vector. Because this encoder used a fixed-length vector, the architecture struggled with longer sequences. This was resolved by the \textit{attention mechanism}~\cite{bahdanau_neural_2016}. In essence, with the attention mechanism the model learns which tokens are relevant for the translation task. With this approach, the attention mechanism improved the performance on longer sequences. Attention got in the spotlight with the transformer architecture, which significantly outperformed the state-of-the-art models~\cite{vaswani_attention_2017}. Where an RNN would learn dependencies between words based on the sequence, a transformer learns dependencies based on the word itself. The transformer architecture eliminated the need for RNNs, making the transformer superior to architectures as ELMo in both prediction accuracy as well as training time~\cite{vaswani_attention_2017}.

\begin{figure}
  \centering
  \includegraphics[scale=0.3]{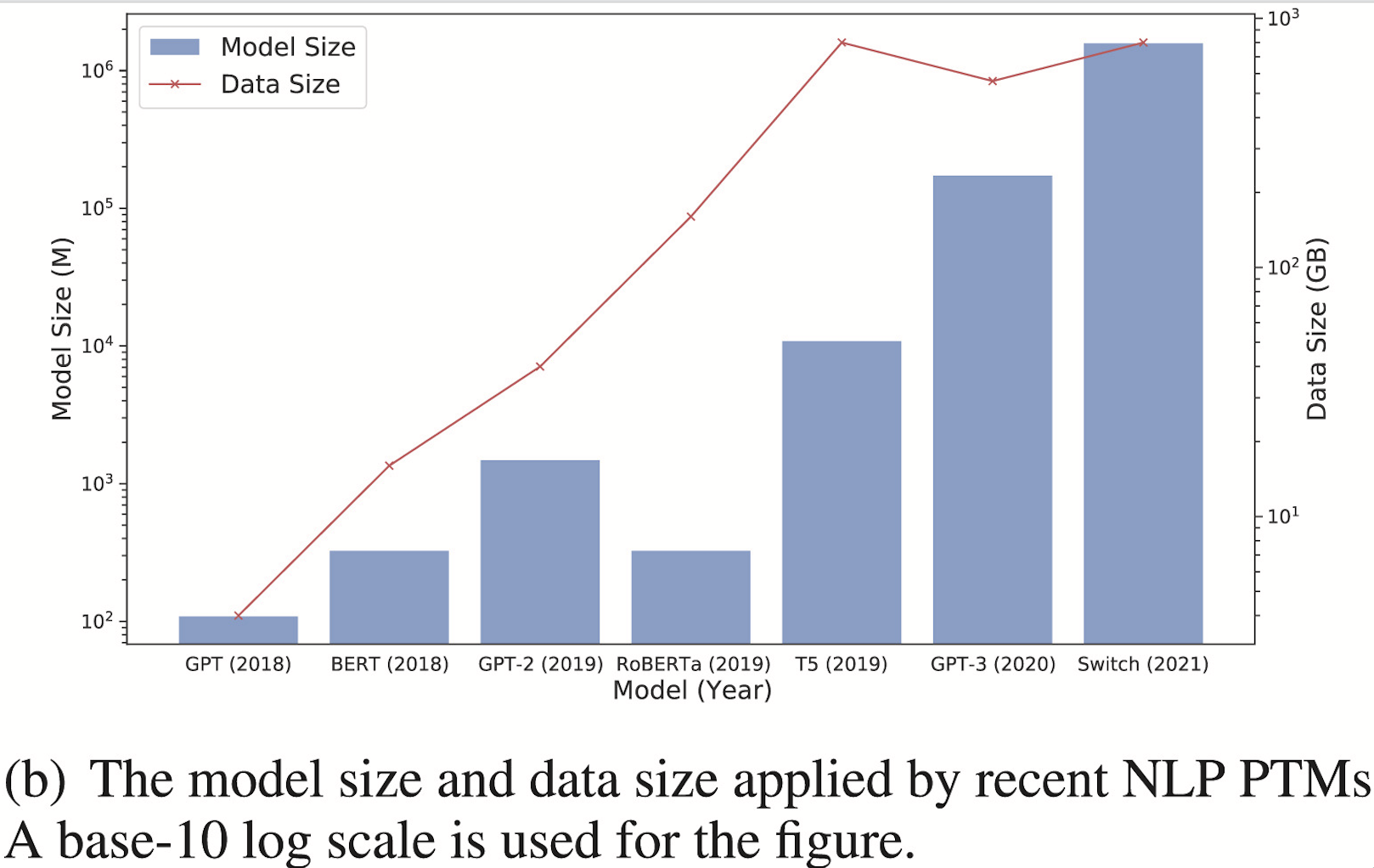}
  \caption{Graph depicting the logarithmic growth of LLMs since GPT-1 in 2018 until 2021. Obtained from~\cite{han_pre-trained_2021}.}
  \label{fig_model_size}
\end{figure}
With GPT-1, released in 2018 having 'just' 117 million parameters, models and datasize grow over time. As depicted in Figure~\ref{fig_model_size}, the growing logarithmic scale shows a clear trend on larger and larger language models. Though this trend seemed to reach a limit with the release of GPT-3's 175 billion parameters (2020), Switch released the biggest LLM so far with 1.57 trillion (1.57 E12) parameters in 2021. However, this is uncertain as developers of GPT-4 do not disclose model architectures anymore due to "the competitive landscape and the safety implications"~\cite{gpt4_2024}. With these LLMs, accessibility to non-commercial users is at stake because the models are so big they cannot be used without access to powerful GPUs.

\subsection{Quantisation}
To resolve this problem with (too) large models, quantisation reduces the model size by transforming weights from FP32 to lower-bit formats such as Int8 or even FP4. This allows the user to run large models on non-GPU devices such as personal computers or even mobile phones~\cite{almeida_smart_2021}. To increase training speed, 32-bit floating point (FP32) weights are already updated using 16-bit floating point (FP16) gradients, making the training twice as fast compared to using FP32 for gradients~\cite{belkada_gentle_2022}. The performance of a quantised model using QLoRA can match the performance of ChatGPT up to 99.3\%~\cite{dettmers_qlora_2023}. There are several methods to quantise model weights, from uniformly quantising to dynamically quantising weights~\cite{wang2024model}. There is an indication that quantisation makes the model more efficient during inference but this depends on various settings~\cite{wang2024model}. A recent paper \cite{rajput2024benchmarking} analyses five quantisation methods on efficiency using the LLAMA-2-7B model. The authors hint that some methods are optimised for GPU work, whilst others are generic.

\subsection{Pruning}
LLMs have many complex paths to come to their prediction. The multi-headed transformer architecture is an example where we observe an impressive performance on machine translation tasks~\cite{vaswani_attention_2017}. However, analysing why multi-headed architectures work so well is challenging. To examine the effect of specific heads, researchers identified the functionality of the various heads~\cite{voita_analyzing_2019}. The authors found that heads have different roles such as positional heads (attending adjacent tokens), synthetic heads (attending syntactic relationship) and attention to rare words. The authors removed, \textit{pruned}, the heads without specialised roles. Surprisingly, this had minimal impact on model performance: the authors pruned nearly 75\% of the encoder heads and 33\% of the decoder heads ``without any noticeable loss in translation quality"~\cite{voita_analyzing_2019}. Do fewer heads in transformers also lead to a more efficient model? With careful selection of parameters, pruning can reduce the model size without considerable compromise on performance~\cite{gholami_can_2023}.

\subsection{Measuring the energy footprint of code}

\textbf{Hardware-based}. The most obvious method to measure energy consumption is by measuring the difference in energy that goes in and comes out of a device. By measuring the difference during the execution of code, e.g. AI, one can derive the energy consumption. This hardware-based solution, also known as the \textit{power plug method}, is often an expensive investment and inadequate for detailed analysis because background processes are also measured~\cite{khan_rapl_2018}. Nevertheless, specifically for larger setups such as a data centre, the power plug method can be a suitable choice to measure the energy consumption of code.

\textbf{Software-based}. The software approach measures the utilization of specific components such as the Central Processing Unit (CPU) and Graphics Processing Unit (GPU). The utilization percentage of such a component can be factored by the power consumption to estimate the energy consumption. Some papers indicate that software-based methods correlate up to 0.99 with hardware-based methods~\cite{khan_rapl_2018}, others indicate a deviation of 20\%~\cite{cao2020towards}. The estimates are not precise, nevertheless, they provide insight into trends, allowing us to contextualise the results. Two software-based methods are often discussed in related work: pyRAPL and CodeCarbon.

\subsubsection{pyRAPL}
Using Intel's Running Average Power Limit (RAPL) interface, we can estimate the energy consumption of Intel chips~\cite{weaver_measuring_2012}. pyRAPL reads RAPL counters to estimate the energy consumption of Python code~\cite{spirals_pyrapl_2019}, correlating up to 0.99 with hardware measurements~\cite{khan_rapl_2018}. Using pyRAPL over hardware holds two advantages: the costs are lower and the setup allows measuring specific parts of the CPU. A disadvantage is that pyRAPL solely works on Intel CPUs, whereas most AI is trained on GPUs~\cite{lacoste_quantifying_2019}.

\subsubsection{CodeCarbon} \label{SecCodeCarbon}
Another open-source software approach is \hyperlink{https://github.com/mlco2/codecarbon
}{CodeCarbon}, which uses the RAPL~\cite{weaver_measuring_2012} and NVIDIA's System Management Interface (SMI). SMI is a command-line interface used to monitor the NVIDIA GPUs~\cite{nvidia_nvidia_2012}, but can therefore also be used to query the GPU for its energy consumption. Therefore, CodeCarbon can measure the energy consumption of both CPUs and GPUs.

\section{Methodology} \label{SecMet}
We aim to make this work reproducible using open-source models and methods discussed in scientific literature. In this section we describe our method, details and implementation follow in Section~\ref{SecExpSet}.

\subsection{Selecting a model}
StarCoder2~\cite{lozhkov2024starcoder} is an open-source code LLM, generated by collaborating scientists in the \href{https://www.bigcode-project.org/}{BigCode} project. StarCoder2 is transparent about training by mentioning both the computational costs and energy impact. Details on the architecture are provided in the StarCoder2 paper and on \href{https://github.com/bigcode-project/starcoder2}{GitHub}. With StarCoder2's statements on open-source development and research contributions, we pick the StarCoder2-3B and StarCoder2-7B models for our experiments.

\subsection{Evaluating the output}
To test the performance of a code LLM we use the task of code completion. In code completion, the model is prompted with a piece of code, i.e. the documentation of a function. With this description, the model generates the code of this function. 'Regular' LLMs are often evaluated with BLUE scores, which measure how close machine-generated output is to human-generated output. The score rates between 0 and 1 and gets closer to 1 when a machine translation gets closer to a human translation. However, BLUE does not work well with code as BLEU measures the similarity whilst for code, many correct solutions are available for the same problem set. So, instead, we use pass@$k$~\cite{chen_evaluating_2021} which measures the probability that for a given coding query, at least one of the top $k$ outputs passes all unit tests.

\subsection{Evaluation framework}
\begin{figure}[h]
    \centering
    \includegraphics[width=\linewidth]{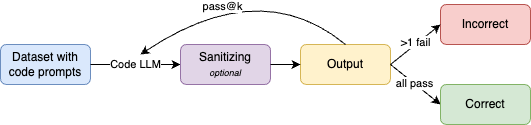}
    \caption{Overview of our evaluation framework. Prompts are fed into the code LLM, the output is then pre-processed after which the output is evaluated. The top-$k$ outputs are and if at least one implementation passes, the pass@$k$ score is 1.}
    \label{FigEvalPlus}
\end{figure}

While some papers draft large data sets to test their models, this does not guarantee that these tests are accurate. Rather, we prefer having a smaller, but very accurate test. As such, we use EvalPlus~\cite{liu_is_2023} as a framework to evaluate the generated code. After preprocessing, the output is then subject to various unit tests that define the right data structures and outputs. If the code passes all of these tests, the output is labeled as correct. For a failure on one of these tests, the output is labeled incorrect. In the end, EvalPlus gives a pass@k score that depicts the mean success rate. A schematic overview of this process is depicted in Figure~\ref{FigEvalPlus}.

\subsection{Quantising model weights}
At least 8 methods compress model weights into smaller datatypes and are supported by the Hugging Face transformer library. \href{https://huggingface.co/docs/transformers/main/quantization/overview}{Hugging Face} provides an overview of which method to pick in what use case. For example, of these 8 methods, 3 methods are not available 'on the fly', meaning that the models need to be calibrated after quantisation. Depending on the hardware, quantisation can take 5 minutes for a 350M parameter model but up to 4 hours to quantise a 175B parameter model. All methods are supported by transformers, but bytsandbytes (BNB) is the most accessible option so we stick to using BNB as quantisation method in our exploration.

\subsection{Pruning the model}
We prune the last layers of StarCoder2-3B and StarCoder2-7B until the pass@1 metric reaches 0. As pruning does not necessarily lead to reduced performance~\cite{gholami_can_2023}, we reduce the model by pruning the last layers as these often contain context where earlier layers encode knowledge~\cite{ju_how_2024}. With each pruned layer we evaluate the compressed models on pass@1 accuracy, inference time and energy consumption. To contextualise the results we will prune the last layers of Phi-2~\cite{javaheripi2023phi} as well and compare the results with our StarCoder2 models.

\section{Experimental setup} \label{SecExpSet}
To reproduction of our experiments, we provide our system specifications below and our implementation on \href{https://github.com/ana-oprescu/GreenLLMs}{GitHub}.

We use the national supercomputer facilities from SURF, and specifically its \href{https://www.surf.nl/en/services/snellius-the-national-supercomputer}{Snellius} environment.

\begin{table}[h]
    \centering
    \begin{tabular}{@{}ll@{}}
        \toprule
        Operating system & Linux-4.18.0 \\
        CPU & \begin{tabular}[c]{@{}l@{}}18x Intel(R) Xeon(R) \\ Platinum 8360Y CPU @ 2.40GHz\end{tabular} \\
        GPU & 1x NVIDIA A100-SXM4-40GB \\ \bottomrule
    \end{tabular}
    \caption{Details of the hardware and operating system.}
    \label{TabSysSpecs}
\end{table}

\section{Results} \label{SecRes}
We first present the results of our experiment where we apply quantisation to StarCoder2-3B and StarCoder2-7B. Then, we show the energy consumption and pass@k after we pruned the last layers of StarCoder2-3B and StarCoder2-7B.

\subsection{Impact of quantisation}
Table~\ref{StarCoder3B7B_pass1} presents the pass@1 score for the StarCoder2-3B and StarCoder2-7B models. The StarCoder2 paper reports a pass@1 score of 31.7 on HumanEval and 27.4 on HumanEval+ for StarCoder2-3B. For StarCoder2-7B, the pass@1 scores are 35.4 and 29.9 respectively. Our results for StarCoder2-3B are thus significantly lower. We verified our setup with \href{https://huggingface.co/microsoft/phi-2}{Phi-2} and saw no difference between our results and the results reported in the Phi-2 paper. We can thus validate our setup and dive into the difference \href{https://github.com/ana-oprescu/GreenLLMs?tab=readme-ov-file#starcoder2-performance}{on GitHub}.

\begin{table}[h!]
\centering
\begin{tabular}{@{}llcc@{}}
\hline
& & \textbf{\begin{tabular}[c]{@{}c@{}}HumanEval\\ (sanitized)\end{tabular}} & \textbf{\begin{tabular}[c]{@{}c@{}}HumanEval+\\ (sanitized)\end{tabular}} \\ \hline
\textbf{3B} & 128 tokens & 8.5 & 7.9 \\
& 256 tokens & \textbf{13.4} & \textbf{10.4} \\
\textbf{7B} & 128 tokens & 4.3 & 3.7 \\ 
& 256 tokens & 7.9 & 6.1 \\ \hline
\end{tabular}
\caption{Overview of the pass@1 score of the StarCoder2-3B and StarCoder2-7B model for two settings with maximum parameters. The rows show the accuracy for the maximum number of tokens the model used, the highest accuracy is in \textbf{bold}.}
\label{StarCoder3B7B_pass1}
\end{table}

\textbf{Pass@1 of quantised StarCoder2 models}\\
Following the pass@1 scores for the original StarCoder2 models, we present the pass@1 scores for the quantised StarCoder2 models in Table~\ref{QuantisedPass}. We show the pass@1 scores for all configurations in tokens and quantisation. In brackets, we report the difference between the quantised models and the full-sized models from Table~\ref{StarCoder3B7B_pass1}.

\begin{table}[h!]
\centering
\begin{tabular}{@{}llcc@{}}
\toprule
 &  & \textbf{\begin{tabular}[c]{@{}c@{}}HumanEval\\ (sanitized)\end{tabular}} & \textbf{\begin{tabular}[c]{@{}c@{}}HumanEval+\\ (sanitized)\end{tabular}} \\ \midrule
\textbf{3B, 128 tokens} & 4bit & 6.1 {\scriptsize (-2.4)} & 6.1 {\scriptsize (-1.8)}\\
 & 8bit & 6.7 {\scriptsize (-1.8)} & 6.7 {\scriptsize (-1.2)}\\ \addlinespace
\textbf{3B, 256 tokens} & 4bit & 12.2 {\scriptsize (-1.2)}& \textbf{10.4} {\scriptsize (0.0)}\\
 & 8bit & 11.0 {\scriptsize (-2.4)} & 9.1 {\scriptsize (-1.3)}\\ \addlinespace
\textbf{7B, 128 tokens} & 4bit & 3.0 {\scriptsize (-1.3)} & 2.4 {\scriptsize (-1.3)}\\
 & 8bit & \textbf{4.3} {\scriptsize (0.0)} & \textbf{3.7} {\scriptsize (0.0)}\\ \addlinespace
\textbf{7B, 256 tokens} & 4bit & 6.7 {\scriptsize (-1.2)}& 4.3 {\scriptsize (-1.8)}\\
 & 8bit & 7.3 {\scriptsize (-0.6)} & \textbf{6.1} {\scriptsize (0.0)}\\ \bottomrule
\end{tabular}
\caption{Pass@1 scores for the various configurations of the quantised StarCoder2-3B and StarCoder2-7B. The tokens refer to the amount of new tokens generated, the bit refers to the quantisation configuration. In \textbf{bold} are the pass@1 scores matching the results from the full-sized counterparts in Table~\ref{StarCoder3B7B_pass1}, the brackets show the deviation in pass@1 score between the original and quantised models.}
\label{QuantisedPass}
\end{table}

\subsection*{Energy consumption of StarCoder2-3B}
\begin{figure}[h!]
    \centering
    \includegraphics[scale=0.33]{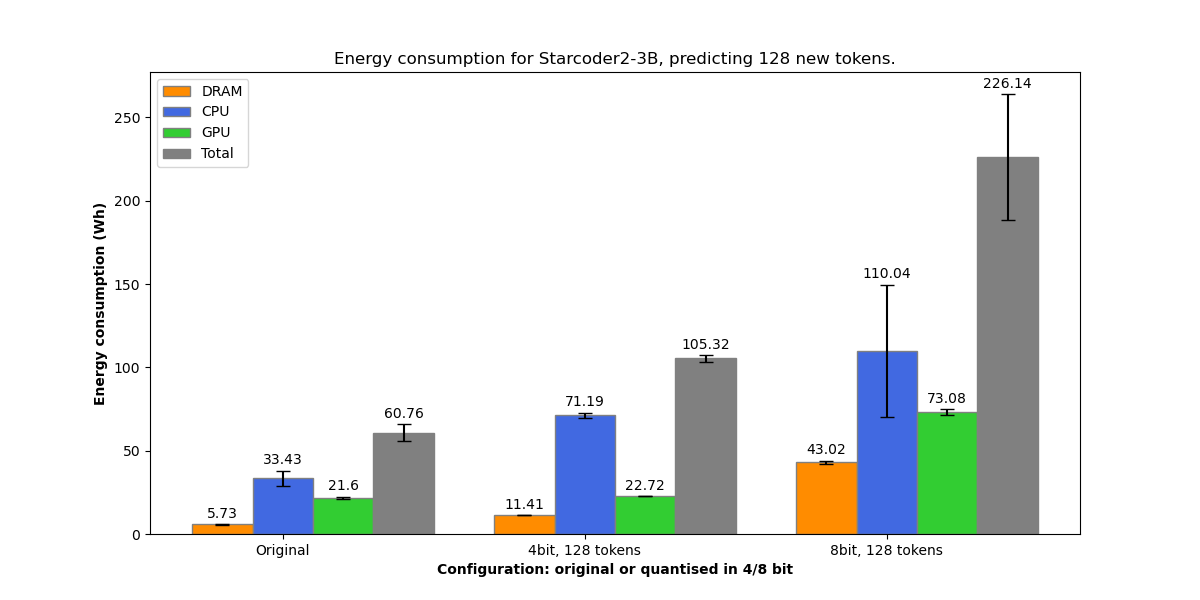}
    \caption{Average energy consumption of the StarCoder2-3B model on five runs, predicting 128 new tokens. The bars indicate the original model and the quantised versions of 4-bit and 8-bit. The whiskers display the 95\% confidence interval (1.96 $\cdot$ std. dev.).}
    \label{Energy_3B_128}
\end{figure}

\begin{figure}[h!]
    \centering
    \includegraphics[scale=0.33]{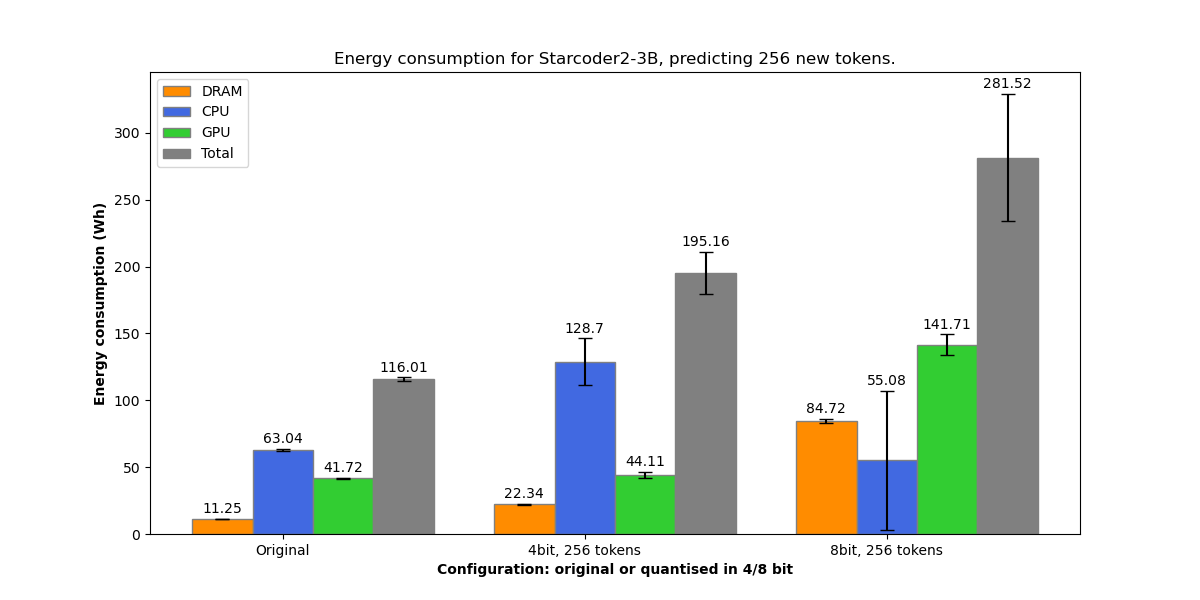}
    \caption{Average energy consumption of the StarCoder2-3B model on five runs, predicting 256 new tokens. The bars indicate the original model and the quantised versions of 4-bit and 8-bit. The whiskers display the 95\% confidence interval (1.96 $\cdot$ std. dev.).}
    \label{Energy_3B_256}
\end{figure}

The energy consumption of StarCoder2-3B predicting 128 tokens is shown in Figure~\ref{Energy_3B_128}, Figure~\ref{Energy_3B_256} for 256 tokens. We interpret and discuss the results in the discussion, Section~\ref{SecDis}.

\subsection*{Energy consumption of StarCoder2-7B}
\begin{figure}[h!]
    \centering
    \includegraphics[scale=0.33]{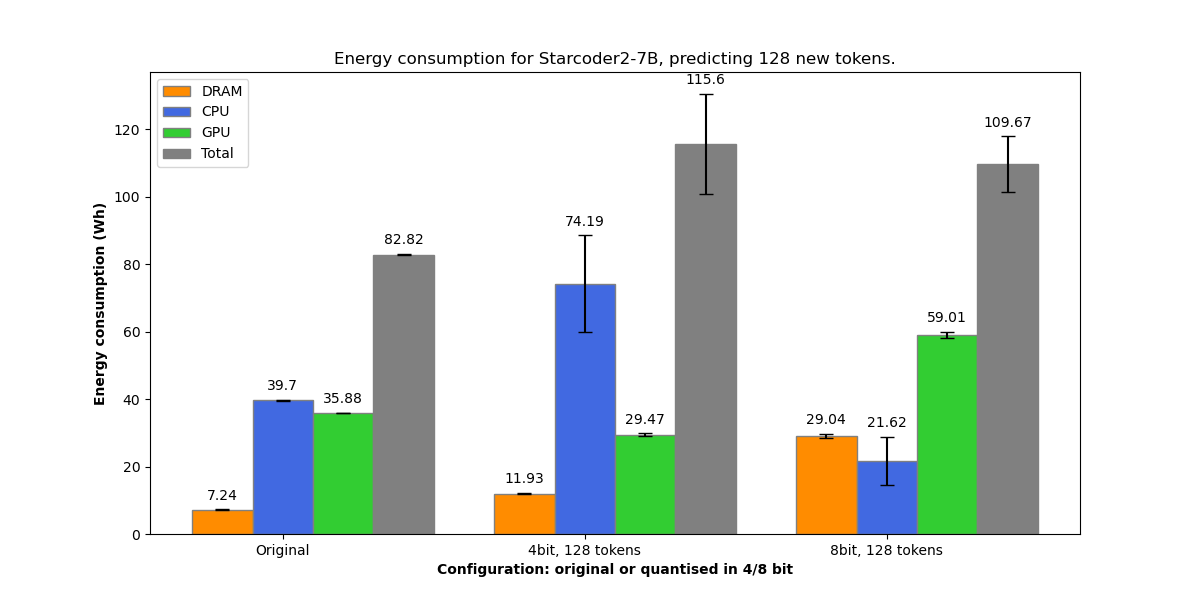}
    \caption{Average energy consumption of the StarCoder2-7B model on five runs, predicting 128 new tokens. The bars indicate the original model and the quantised versions of 4-bit and 8-bit. The whiskers display the 95\% confidence interval (1.96 $\cdot$ std. dev.).}
    \label{Energy_7B_128}
\end{figure}

\begin{figure}[h!]
    \centering
    \includegraphics[scale=0.33]{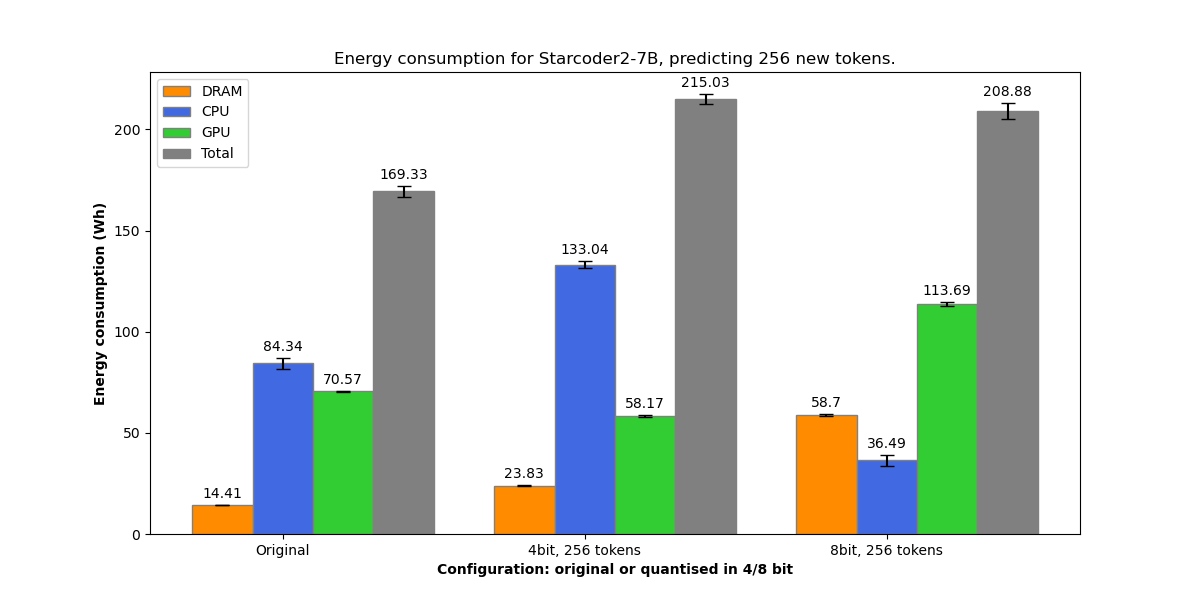}
    \caption{Average energy consumption of the StarCoder2-7B model on five runs, predicting 256 new tokens. The bars indicate the original model and the quantised versions of 4-bit and 8-bit. The whiskers display the 95\% confidence interval (1.96 $\cdot$ std. dev.).}
    \label{Energy_7B_256}
\end{figure}

The energy consumption of StarCoder2-7B predicting 128 tokens is shown in Figure~\ref{Energy_7B_128}, Figure~\ref{Energy_7B_256} for 256 tokens. We interpret and discuss the results in the discussion, Section~\ref{SecDis}.

\subsection*{Runtime of the experiments}
Figures~\ref{Energy_3B_128}-\ref{Energy_7B_256} show that for each model with quantisation, the energy consumption increased. During our experiments we noticed that quantised experiments required more time, which is why we report the runtime as well. Table~\ref{AverageExecTime2} shows the average execution times and throughput (number of tokens predicted per second) of the experiments.

\begin{table}[h!]
\centering
\begin{tabular}{@{}llccc@{}}
\toprule
 &  & \textbf{\begin{tabular}[c]{@{}c@{}}Avg. time\\ (seconds)\end{tabular}} & \textbf{\begin{tabular}[c]{@{}c@{}}Std. deviation\\ (seconds)\end{tabular}} & \textbf{\begin{tabular}[c]{@{}c@{}}Throughput\\ (\# tokens per second)\end{tabular}} \\ \midrule
\textbf{3B, 128 tokens} & Original & 458.3 & 3.6 & 45.8 \\
 & 4-bit & 908.3 & 10.8 & 23.1 \\
 & 8-bit & 3441.2 & 40.1 & 6.1 \\ \addlinespace
\textbf{3B, 256 tokens} & Original & 900.0 & 6.9 & 46.6 \\
 & 4-bit & 1787.3 & 15.9 & 23.5 \\
 & 8-bit & 6766.2 & 55.2 & 6.2 \\ \addlinespace
\textbf{7B, 128 tokens} & Original & 579.3 & 1.0 & 36.2 \\
 & 4-bit & 954.4 & 8.2 & 22.0 \\
 & 8-bit & 2323.3 & 21.6 & 9.0 \\ \addlinespace
\textbf{7B, 256 tokens} & Original & 1152.9 & 2.2 & 36.4 \\
 & 4-bit & 1906.0 & 12.4 & 22.0 \\
 & 8-bit & 4696.0 & 19.7 & 8.9 \\ \bottomrule
\end{tabular}
\caption{Execution time for various configurations of the StarCoder2-3B and StarCoder2-7B models. Each model predicted either 128 or 256 tokens for the 164 tasks in the HumanEval+ set. We show the average execution time over these five experiments, the standard deviation to this average, and the average number of tokens per second (throughput).}
\label{AverageExecTime2}
\end{table}

\subsection{Impact of pruning}
\begin{figure}[h!]
    \centering
    \includegraphics[scale=0.55]{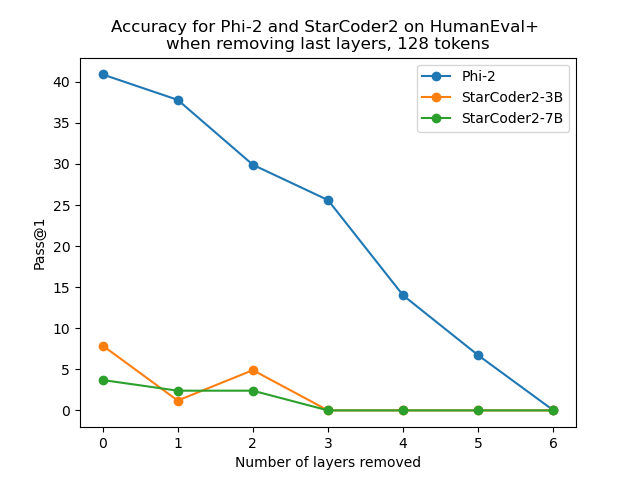}
    \caption{This graph shows the pass@1 score for Phi-2, StarCoder2-3B and StarCoder2-7B on the HumanEval+ benchmark. Each model predicted a maximum of 128 new tokens.}
    \label{PassPruning}
\end{figure}

\begin{figure}[h!]
    \centering
    \includegraphics[scale=0.33]{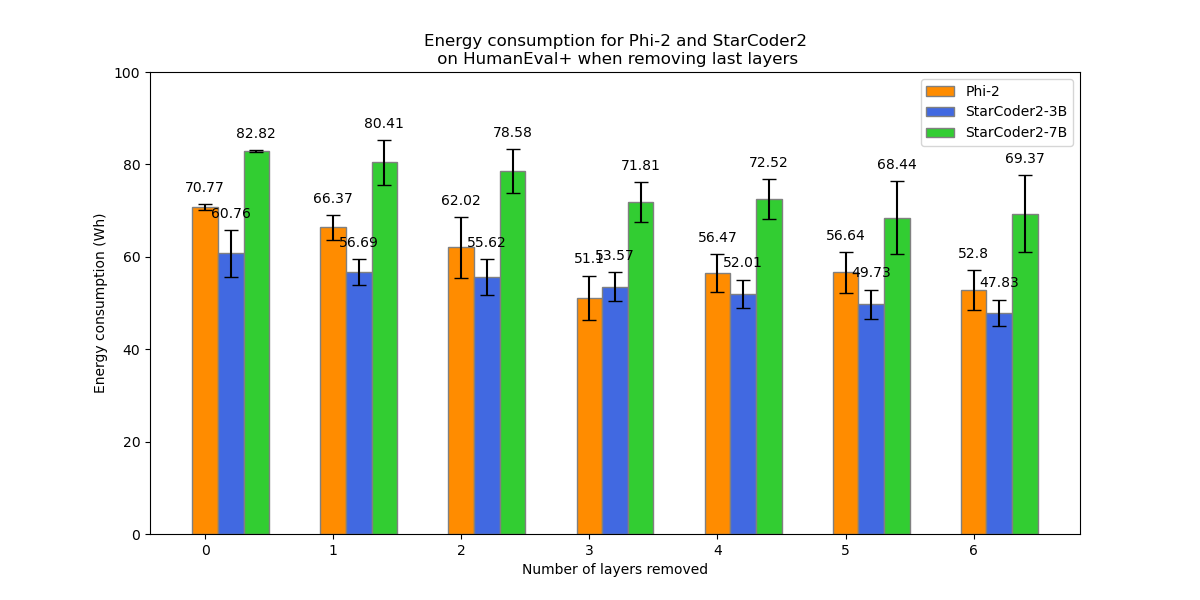}
    \caption{This graph shows the energy consumption for Phi-2, StarCoder2-3B and StarCoder2-7B for each layer. Each model predicted a maximum of 128 new tokens and experiments were done fivefold. The whiskers display the 95\% confidence interval (1.96 $\cdot$ std. dev.)}
    \label{EnergyPruning}
\end{figure}

We analyse the impact of pruning of the last layers by examining the pass@1 score on HumanEval+ and the energy consumption. Figure~\ref{PassPruning} shows the pass@1 score for Phi-2, StarCoder2-3B and StarCoder2-7B. At 0 layers removed the performance equals that of the original, full-sized model. The layers are removed 1-by-1 until the pass@1 score reaches the floor of 0. The energy consumption of this experiment is provided in Figure~\ref{EnergyPruning}. We interpret and discuss the results in the discussion, Section~\ref{SecDis}.

\section{Discussion} \label{SecDis}
Table~\ref{QuantisedPass} shows a clear deviation from pass@1 scores compared to the reported scores in the StarCoder2 paper~\cite{lozhkov2024starcoder}. We discuss this in \href{https://github.com/ana-oprescu/GreenLLMs/edit/main/README.md}{our repository}, but as we are interested in the impact of model compression we use the results as given.

\subsection{Quantisation}
\subsubsection*{\textbf{pass@1}}
Table~\ref{QuantisedPass} shows the pass@1 scores for the quantised models and the deviation between these quantised scores and those of the full-sized models. First, we examine \textbf{StarCoder2-3B} on 128 tokens. In Table~\ref{StarCoder3B7B_pass1}, we see that for both 4-bit quantised models, the pass@1 score is 6.1, 1.8 points lower $\pm -25\%$, than the pass@1 of the original model (7.9). For the 8-bit configuration, the pass@1 score is 6.7, 1.2 points lower $\pm -20 \%$ than the original model (7.9). These reductions are greater than we would expect based on related work~\cite{dettmers_8-bit_2022}. However, if we proceed to the same StarCoder2-3B model predicting 256 tokens, we see that the pass@1 score for the 4-bit configuration matches the pass@1 score of 10.4 from the original model, in line with related work~\cite{dettmers_8-bit_2022}. The 8-bit configuration is 1.3 points lower than the original pass@1 of 10.4. \textbf{StarCoder2-7B} shows a similar trend for 128 tokens, where the 4-bit configuration results in a pass@1 score that is 1.3 lower ($\pm$ -33\%) than the original (3.7). For the 8-bit configuration, we see that the pass@1 score matches the original model (3.7), following related work~\cite{dettmers_8-bit_2022}. For 256 tokens, the 4-bit configuration yields a 1.8 points lower pass@1 ($\pm$ -25\%) score but the 8-bit configuration again matches the pass@1 score of 6.1 for HumanEval+.

\subsubsection*{\textbf{Energy consumption}}
For all 8-bit quantised models in Figures~\ref{Energy_3B_128}-\ref{Energy_7B_256}, we see high confidence intervals, indicating variance in the energy measurements. As the Snellius cluster is a shared computing environment where we used $\frac{1}{4}^{\text{th}}$ of a node, other users might impact the frequency of the CPU or GPU which could lead to errors. We reran the models with the \href{https://gitlab.bsc.es/ear_team/ear/-/wikis/home}{Energy Aware Runtime (EAR)}\footnote{EAR is an energy management framework for high-performance computing. The documentation of EAR is available \href{https://www.bsc.es/sites/default/files/public/bscw2/content/software-app/technical-documentation/ear_guide.pdf}{online}} flag. Our analysis shows that the average CPU and GPU frequency did not change during our experiments, ruling out the hypothesis that other users on the same node impact the frequency. Next to that Figures~\ref{Energy_3B_128}-\ref{Energy_7B_256} all show very little variance for the original models, the variation is mainly tied to the quantisation strategy. The bitsandbytes authors report no instabilities for 8-bit optimisers~\cite{dettmers_8-bit_2022}, but examined training and not the inference.

We observe that quantisation leads to increased energy consumption, regardless of the model or the number of tokens. Our data shows that 8-bit quantisation leads to the highest increase, from 19\%-75\%. 4-bit quantisation seems more efficient and increases by 19\%-43\%. As we expected a reduction in energy consumption, this result was unexpected. Table~\ref{AverageExecTime2} clarifies the higher energy consumption of quantisation because quantised models require more time. For StarCoder2-3B the throughput halves from $\pm$ 46 to $\pm$ 23 in 4-bit quantisation and reduces $\times \frac{1}{7}$, from $\pm$ 46 to $\pm$ 6, for 8-bit quantisation regardless of the number of tokens predicted. However, we see that the energy consumption from Figure~\ref{Energy_3B_128} and Figure~\ref{Energy_3B_256} does not double. So, the energy consumption did not increase proportional to the increased runtime. The energy required per token is lower but this reduction is compensated by a longer inference. For StarCoder2-7B, the reduction in throughput is $\times \frac{2}{3}$, from $\pm$ 36 to $\pm$ 22, for 4-bit quantisation and $\times \frac{1}{4}$, from $\pm$ 36 to $\pm$ 9, for 8-bit quantisation. The throughput is again lower for quantised models, but the reduction is less compared to StarCoder2-3B. A recent paper shows that the bitsandbytes module has a very low throughput compared to other quantisation methods~\cite{rajput2024benchmarking}. This paper strengthens our findings that the quantisation method negatively impacts energy consumption. Related work also describes increased inference time for bitsandbytes~\cite{argerich_measuring_2024}. Even though the energy consumption per second is lower for the quantised models, the throughput is lower than for the original model. Ultimately, the model might spend less energy per second but if the time increases so does the energy consumption. Likely, this is the result of table look-ups during dequantisation. To the best of our knowledge, no literature explains why this method is slower than the original model. Bitsandbytes does not linearly quantise weights, when the model predicts its next token, it has to dequantise the weights again but because the scale is different we do so for each new token. We hypothesized that this might cause cache misses in memory, which can increase the runtime of code~\cite{ghosh_cache_1997}. Our analysis confirms that L1 and L2 cache misses increase with quantised models, we provide these results \href{https://github.com/ana-oprescu/GreenLLMs/edit/main/README.md}{on GitHub}. Next to cache misses, GPUs are not created to accelerate matrix multiplication of integers but rather for floating point numbers~\cite{guo_fast_2024, patel_characterizing_2024, wang2024model}.

Given these results from Tables~\ref{StarCoder3B7B_pass1} and~\ref{QuantisedPass}, there are two interesting findings. First, we empirically show that in some cases the accuracy for quantised models can match that of original models, reinforcing the claims from~\cite{dettmers_8-bit_2022}. Second, the StarCoder2-7B seems to perform best with 8-bit quantisation whereas for StarCoder2-3B this is inconclusive as 4-bit or 8-bit does not consistently outperform its counterpart. Energy-wise, quantisation consistently increases energy consumption.


\subsection{Pruning}
Figure~\ref{PassPruning} shows the pass@1 score for StarCoder2-3B, StarCoder2-7B and Phi-2 for various pruned layers when predicting 128 tokens. We see a clear downward trend, indicating that pruning layers negatively impact the pass@1 score. For Phi-2 the pass@1 score decreases by around 15\% per removed layer. When more layers are pruned, the pass@1 score decreases accordingly except for the pruning of the last two layers of StarCoder2-3B, which slightly improves before decreasing again. When we prune 3 layers, both StarCoder2 models become erroneous and fail all tasks from HumanEval+. StarCoder2-3B has 30 layers and StarCoder2-7B has 32 layers. Interestingly, Phi-2 shows a rather consistent decline and with 3 pruned layers Phi-2 still outperforms the original StarCoder2 models. Phi-2 has 31 layers, similar to the StarCoder2 model, but nevertheless seems a better fit to predict without its final layers. We suspect that the instruction-tuning of Phi-2 plays a role in the higher pass@1 scores. To indicate why the models become less accurate we had a closer look at the outcomes of the individual tasks. This shows that the more layers we prune, the more the model repeats itself. This holds for both StarCoder2 models and for Phi-2. Literature suggests that removing the final layers has a more significant impact than middle layers~\cite{ma_llm-pruner_2023}, by pruning and retraining this reduction can be minimised. We did not retrain after pruning, we suggest this in our future work.

Figure~\ref{EnergyPruning} shows the energy consumption of Experiment 3. Apart from the large confidence interval, we see a trend in which removing layers results in reduced energy consumption for all models. This reduction meets expectations because the model does fewer computations before presenting the output. The data points for all models often lie close together, but outliers shift the average and increase the error whiskers.


\subsection{Threats to validity}
\subsubsection*{CPU frequency}
Using the EAR framework, we found some deviations in the average CPU frequency. With our experimental setup with Slurm as job scheduler, we cannot target individual nodes. Therefore all our experiments were done on different nodes, but for each task a different node was assigned. Figure~\ref{EnergyPruning} was the only experiment done on five different nodes and clearly shows the impact of these deviations in our measurements. The exact values of this thesis hence might be biased, but we are confident that our reported trends will hold across other setups as well. Our results show that some nodes lead to consistent outliers over measurements. If we could target one node this error would be systematic and not introduce variance in comparisons. Regardless of this internal limitation and variance, our experiments show clear trends that answer our research questions. 

\subsubsection*{CodeCarbon}
During the experiments, the framework CodeCarbon sometimes gave a warning that the \textit{'Background scheduler didn't run for a long period, results might be inaccurate'}. This warning was also reported as an \href{https://github.com/connectomicslab/connectomemapper3/discussions/221}{issue on GitHub}. Though the reported issue also contained messages that some jobs were missed, which did not occur in our experiment, this implies that parallelisation could impact the accuracy of measurements. We did not compare different parallelisations in the same experiment we deem this to have no to very little impact.

\subsubsection*{Optimising techniques}
We did not use techniques such as multi-threading or parallelization in our experiments. These techniques can be beneficial for sequential tasks such as next-token prediction in LLMs. Our limited time did not allow for this analysis but future work should expand this work by deploying these methods to find if our conclusions hold.

\subsection{Future work}
\underline{RD1: Dedicated hardware}.
\cite{wang2024model} report that ``Quantisation is the most straightforward method to cut down memory cost and improve inference speed for LLMs, especially on hardware with support for fast operation of low-bit datatypes". As our results indicate that the energy consumption per second is lower for quantised models, using dedicated hardware could reduce runtime and consequently energy consumption. 


\underline{RD2: Expand to other models}. Our work in an exploration of StarCoder2, but LLMs are released by the hour. Future work should look into the effect of quantisation on other LLMs to see if we can generalise the findings.

\underline{RD3: Targetted pruning}. We use a naive pruning method to incrementally prune the last layer. More advanced pruning strategies could however lead to a more balanced trade-off between size and performance~\cite{gholami_can_2023}.

\section{Related work} \label{SecRelWork}
\subsection{Throughput of various quantisation methods}
A recent paper by Rajput (2024)~\cite{rajput2024benchmarking} analyses five quantisation methods on LLAMA-2-7B, amongst which Bits and Bytes (BNB). The authors find that BNB is the worst-performing quantisation method in efficiency and perplexity. The authors hint that the best method is optimised for GPU work, compared to BNB which is generic. The authors call for energy optimization an explicit design criterion alongside accuracy. The authors use perplexity on the  wikitext-2 dataset. Our work differs from this work as it focuses on coding tasks and studies the StarCoder2 models instead of LLAMA-2.


\subsection{Estimating the carbon footprint of BLOOM}
The authors estimate the Carbon Footprint of BLOOM, a 176B parameter LLM~\cite{luccioni_estimating_2023}. The authors simulated training and inference, and convert the energy used in~\coo-emissions. This is one of the first papers that assesses the LCA of a LLM, nevertheless, it underestimates the impact of inference. Our work differs as we try to reduce the energy consumption during inference, this work estimates the carbon footprint of BLOOM's lifecycle. 

\subsection{Model Compression and Efficient Inference for LLMs}
This paper surveys the area of model compression and its effect on efficient inference~\cite{wang2024model}. The paper describes pruning, quantisation and knowledge distillation and compares the advantages and disadvantages of each method. Our work differs as we conduct experiments with easy-to-use methods to explore how we can reduce the energy consumption.



\section{Conclusion} \label{SecCon}
More than 100 million people interact with ChatGPT and programmers using code LLMs describe increased productivity. But this mass adoption comes at a cost: AI consumes a significant amount of electricity, causing the \coo-emissions to soare. With global temperatures now 1.48 degrees Celsius above the pre-industrial age, irreversible climate change becomes more likely unless we reduce greenhouse gas emissions. To contribute to this reduction in greenhouse gasses, we construct three experiments to answer our research question: \textbf{How can we reduce the energy consumption of code Large Language Models with minimal harm to accuracy?} We use quantisation to compress the weights of the 3B and 7B StarCoder2 models. We evaluated these quantised models on HumanEval+ and found that the pass@1 score for StarCoder2-7B with 8-bit quantisation matches the full-sized model. For all other configurations, the pass@1 score on HumanEval+ is lower than their full-sized counterparts. The energy consumption increased by at least 19\% for all configurations, presumably because quantised models have a lower throughput. However, the reduced throughput is not proportional to the energy consumption. This shows that the average amount of energy per second was reduced, but the longer execution times surpass this reduction. In our second experiment, we pruned the last layers of Phi-2, StarCoder2-3B and StarCoder2-7B. We found that the performance dropped significantly with the number of layers that were pruned, whereas the energy reduction was minimal. We therefore deem the pruning of layers unsuitable for reducing the energy consumption of code LLMs. For future work, we deem an improved throughput for quantised models a promising field to reduce the energy consumption of code LLM inference.


\bibliography{references}
\bibliographystyle{IEEEtran}

\end{document}